\documentclass[11pt,drafunctiontcls,onecolumn]{IEEEtran}

\usepackage{	latexsym,%
		    	amsmath,%
			amssymb,%
			amsthm,
			graphicx,
			relsize,
			caption,
			accents,
			caption,
			subcaption,
			tikz,
			pgfplots,
			bm,
			mathtools}

\usepackage[section]{placeins}
\usepackage[inline]{enumitem}
\usepackage[utf8]{inputenc}
\usepackage{array}
\usepackage{cite}
\usepackage[ruled,vlined]{algorithm2e}

\pgfplotsset{compat=1.5.1}
\usetikzlibrary{patterns}
\usetikzlibrary{shapes.geometric}
\usetikzlibrary{shapes,snakes,patterns}
\usetikzlibrary{arrows,positioning,automata,calc}
\usetikzlibrary{plotmarks}

\usepackage{microtype}
\usepackage{graphicx}
\usepackage{booktabs} %

\usepackage{hyperref}

\definecolor{bleudefrance}{rgb}{0.19, 0.55, 0.91}

\newcommand{\R}{\mathbb{R}}
\newcommand{\N}{\mathbb{N}}

\newcommand{\set}[1]{\left\{#1\right\}}
\newcommand{\indicator}[1]{1_{\set{#1}}}
\newcommand{\E}{\operatorname{\mathbb{E}}}
\newcommand{\ip}[2]{\ensuremath{\left\langle #1,#2\right\rangle}}
\newcommand{\Var}{\operatorname{Var}}

\newcommand{\norm}[1]{\left\|#1\right\|}
\newcommand{\ghat}{\hat{\mathbf{g}}}
\newcommand{\bw}{\mathbf{w}}
\newcommand{\by}{\mathbf{y}}

\newcolumntype{L}[1]{>{\raggedright\let\newline\\\arraybackslash\hspace{0pt}}m{#1}}
\newcolumntype{C}[1]{>{\centering\let\newline\\\arraybackslash\hspace{0pt}}m{#1}}
\newcolumntype{R}[1]{>{\raggedleft\let\newline\\\arraybackslash\hspace{0pt}}m{#1}}

\newcommand{\abs}[1]{\left\lvert #1\right\rvert}

\renewcommand{\ge}{\geqslant}
\renewcommand{\le}{\leqslant}

\newtheorem{theorem}{Theorem}
\newtheorem{example}{Example}
\allowdisplaybreaks[2]

\newtheorem{lemma}{Lemma}

\newtheorem{proposition}{Proposition}

\newtheorem{remark}{Remark}

\begin{document}

\title{Adaptive Stochastic Gradient Descent for Fast and Communication-Efficient Distributed Learning}
\author{

\IEEEauthorblockN{Serge Kas Hanna\IEEEauthorrefmark{1}, Rawad Bitar\IEEEauthorrefmark{1}, Parimal Parag\IEEEauthorrefmark{2}, Venkat Dasari \IEEEauthorrefmark{3}, and  Salim El Rouayheb\IEEEauthorrefmark{4} }

\IEEEauthorblockA{ \small{
\IEEEauthorrefmark{1} Institute for Communications Engineering, Technical University of Munich, Germany \\
\IEEEauthorrefmark{2} Department of Electrical Communication Engineering, Indian Institute of Science, Bengaluru, KA, India  \\
\IEEEauthorrefmark{3} US Army Research Laboratory, Aberdeen Proving Ground, MD, USA  \\
\IEEEauthorrefmark{4} Department of Electrical and Computer Engineering, Rutgers University, Piscataway, NJ, USA \\
{\footnotesize Emails: \{serge.k.hanna, rawad.bitar\}@tum.edu, parimal@iisc.ac.in, venkateswara.r.dasari.civ@mail.mil, salim.elrouayheb@rutgers.edu}
} 
\thanks{This paper was presented in part at the 2020 IEEE International Conference on Acoustics, Speech and Signal Processing (ICASSP)~\cite{ICASSP}.}
}
}
\maketitle
\vspace{-1cm}
\begin{abstract}
We consider the setting where a master wants to run a distributed stochastic gradient descent (SGD) algorithm on $n$ workers, each having a subset of the data. Distributed SGD may suffer from the effect of stragglers, i.e., slow or unresponsive workers who cause delays. One solution studied in the literature is to wait at each iteration for the responses of the fastest $k<n$ workers before updating the model, where $k$ is a fixed parameter. The choice of the value of $k$ presents a trade-off between the runtime (i.e., convergence rate) of SGD and the error of the model. Towards optimizing the error-runtime trade-off, we investigate distributed SGD with adaptive~$k$, i.e., varying $k$ throughout the runtime of the algorithm. We first design an adaptive policy for varying $k$ that optimizes this trade-off based on an upper bound on the error as a function of the wall-clock time that we derive. Then, we propose and implement an algorithm for adaptive distributed SGD that is based on a statistical heuristic. Our results show that the adaptive version of distributed SGD can reach lower error values in less time compared to non-adaptive implementations. Moreover, the results also show that the adaptive version is communication-efficient, where the amount of communication required between the master and the workers is less than that of non-adaptive versions. %
\end{abstract}

\section{Introduction}
\label{Intro}
We consider a distributed computation setting in which a master wants to learn a model on a large amount of data in his possession by dividing the computations on $n$ workers. The data at the master consists of a matrix $X\in \R^{m\times d}$ representing $m$ data vectors $\mathbf{x}_\ell$, $\ell = 1,\dots,m$, and a vector $\mathbf{y} \in \R^{m}$ representing the labels of the rows of $X$. Define $A\triangleq [X|\mathbf{y}]$ to be the concatenation of $X$ and $\by$. The master would like to find a model $\bw^\star\in \R^d$ that minimizes a loss function $F(A, \bw)$, i.e,
\begin{equation}
\bw^\star = \arg\min_{\bw \in \R^d} F(A, \bw).
\end{equation}
This optimization problem can be solved using Gradient Descent (GD), which is an iterative algorithm that consists of the following update at each iteration~$j$, 
\begin{equation}\label{eq:update}
\bw_{j+1} = \bw_j -\eta \nabla F(A, \bw_j) \triangleq   \bw_j -  \dfrac{\eta}{m} \sum_{\ell=1}^m \nabla F(\mathbf{a}_\ell, \bw_j) ,
\end{equation}
where $\eta$ is the step size, and $\nabla F(A,\bw)$ is the gradient of $F(A,\bw)$. To distribute the computations, the master partitions the data equally to $n$ workers. The data partitioning is horizontal, i.e., each worker receives a set of rows of $A$ with all their corresponding columns. Let  $S_i$ be the sub-matrix of $A$ sent to worker $i$. Each worker computes a partial gradient defined as
\begin{equation} 
\nabla F(S_i, \bw_j) \triangleq \dfrac{1}{s}\sum_{\mathbf{a}_\ell \in S_i} \nabla F(\mathbf{a}_\ell, \bw_j),
\end{equation}
where $s = m/n$ is the number of rows in $S_i$ (assuming $n$ divides $m$). The master computes the average of the received partial gradients to obtain the actual gradient $\nabla F(A,\bw)$, and then updates $\bw_j$. 

In this setting, waiting for the partial gradients of all the workers slows down the process as the master has to wait for the stragglers \cite{DB13}, i.e., slow or unresponsive workers, in order to update $\bw_j$. Many approaches have been proposed in the literature to alleviate the problem of stragglers. A natural approach is to simply ignore the stragglers and obtain an estimate of the gradient rather than the full gradient, see \cite{Gauri,CPMBJ16}. This framework emerges from single-node (non-distributed) mini-batch stochastic gradient descent (SGD) \cite{RM51}. Batch SGD is a relaxation of GD in which $\bw_j$ is updated based on a subset (batch) $B$ $(|B|<m)$ of data vectors that are chosen uniformly at random at each iteration from the set of all $m$ data vectors, i.e.,
\begin{equation}\label{eq:SGDup}
\bw_{j+1} =  \bw_j - \dfrac{\eta}{|B|}\sum_{\mathbf{a}_\ell \in B} F(\mathbf{a}_\ell, \bw_j).
\end{equation}
It is shown that SGD converges to $\bw^\star$ under mild assumptions on the loss function $F(A,\bw)$, but may require a larger number of iterations as compared to GD \cite{Bottou,RM51,BT89,CSSS11,AD11,DGSX12,SS14}. 

Consider the approach where the master updates the model based on the responses of the fastest $k<n$ workers and ignores the remaining stragglers. Henceforth, we call this approach {\em fastest-k SGD}. The corresponding update rule for fastest-$k$ SGD is given by
\begin{equation}\label{eeq6}
\mathbf{w}_{j+1}=\mathbf{w}_j-\frac{\eta}{k  }  \sum_{i \in R_j} \nabla F(S_{i},\mathbf{w}_j)\triangleq \mathbf{w}_j-\eta ~\ghat(\mathbf{w}_j) ,
\end{equation} 
where $R_j$ is the set of the fastest $k$ workers at iteration~$j$; and $\ghat(\mathbf{w}_j)$ is the average of the partial gradients received by the master at iteration $j$ which is an estimate of the full gradient $\nabla F(A,\mathbf{w}_j)$. Note that if we assume that the response times of the workers are random {\em iid}, then one can easily show that fastest-$k$ SGD is essentially equivalent to single-node batch SGD since the master updates the model at each iteration based on a uniformly random batch of data vectors belonging to the set of the fastest $k$ workers. Therefore, fastest-$k$ SGD converges to $\bw^\star$ under the random {\em iid} assumption on the response times of the workers and standard assumptions on the loss function $F(A,\bw)$.

The convergence rate of distributed SGD depends on two factors simultaneously: \begin{enumerate*}[label=(\roman*)] \item the error in the model versus the number of iterations; \item the time spent per iteration. \end{enumerate*} Therefore, in this work we focus on studying the convergence rate with respect to the wall-clock time rather than the number of iterations. In fastest-$k$ SGD with fixed step size, the choice of the value of $k$ presents a trade-off between the convergence rate and the error floor. Namely, choosing a small value of $k$ will lead to fast convergence since the time spent per iteration would be short, however, this will also result in a low accuracy in the final model, i.e., a higher error floor. Towards optimizing this trade-off, we study adaptive policies for fastest-$k$ SGD where the master starts with waiting for a small number of workers $k$ and then gradually increases $k$ to minimize the error as a function of time. Such an optimal adaptive policy would guarantee that the error is minimized at any time instance. This would be particularly useful in applications where SGD is run with a deadline since we would achieve the best accuracy within any time restriction.
 
\subsection{Related work} 
\subsubsection{Distributed SGD} The works that are closely related to our work are that of \cite{Gauri,CPMBJ16}. In \cite{CPMBJ16} the authors study fastest-$k$ SGD for a predetermined $k$. In \cite{Gauri}, the authors consider the same setting as \cite{CPMBJ16} and analyze the convergence rate of fastest-$k$ SGD with respect to the number of iterations. In addition to the convergence analysis with respect to the number of iterations, the authors in \cite{Gauri} separately analyze the time spent per iteration as a function of $k$. 

Several works proposed distributing the data to the workers redundantly. The master then uses coding theoretical tools to recover the gradient in the presence of a fixed number of stragglers~\cite{TLDK17,YA18,raviv2017gradient,lee2018speeding,ferdinand2018anytime,yu2018lagrange,kiani2018exploitation,chen2018draco,karakus2017straggler,halbawi2017improving,DCG16,KS18,fahim2017optimal,BPR17,Ozfatura2020,kadhe2020communication}. In \cite{charles2017approximate,MRM18,bitar2020stochastic,wang2019erasurehead,wang2019fundamental,glasgow2021approximate,Ozfatura2021}, the authors propose a strategy known as approximate gradient coding in which the master distributes the data redundantly and uses coding techniques to compute an estimate of the gradient. The quality of the estimate of the gradient, and thus the convergence speed, depends on the number of stragglers. The authors of~\cite{egger2022efficient} combine our proposed strategy with the multi-armed bandits framework to allow the master to only use the fastest $k$ workers at each iteration.

Note that the setting of the previously mentioned works, and the setting of interest for our work, focuses on the so-called synchronous SGD in which the workers have the same model at each iteration. The literature also studies the asynchronous setting. In asynchronous distributed SGD, whenever a worker finishes its assigned computation, it sends the result to the master who directly updates $\bw$ and sends an updated $\bw$ to that worker who starts a new computation of the partial gradient while the other workers continue their previous computation~\cite{Gauri,recht2011hogwild,liu2015asynchronous,shalev2013accelerated,reddi2015variance,pan2016cyclades}.

\subsubsection{Single-node SGD} Murata \cite{Murata} showed that irrespective of its convergence speed, the single-node SGD algorithm with fixed step size goes through a transient phase and a stationary phase. In the transient phase, $\bw_j$ approaches $\bw^\star$ exponentially fast in the number of iterations. Whereas, in the stationary phase, $\bw_j$ oscillates around $\bw^\star$. Note that  if a decreasing step size over the iterations is used, then $\bw_j$ converges to $\bw^\star$ rather than oscillating around it, however, this leads to a long transient phase and hence a lower convergence rate. To detect the phase transition, \cite{chicago} uses a statistical test based on Pflug's method \cite{Pflug} for stochastic approximation. Detecting the phase transition serves many purposes, such as indicating when to stop the SGD algorithm or when to start implementing further tricks to reduce the distance between $\bw_j$ and $\bw^\star$. In this paper, we build on this line of work to derive the times at which the master should start waiting for more workers in fastest-$k$ SGD.

In another line of work on single-node SGD, the authors in~\cite{google} suggested increasing the batch size with the number of iterations as an alternative to decreasing the step size. The results in~\cite{google} show that increasing the batch size while keeping a constant step size, leads to near-identical model accuracy as decreasing the step size, but with fewer parameter updates, i.e., a shorter training time.

\subsection{Our contributions}
In our work, we consider distributed systems with scarce communication resources that suffer from the effect of stragglers. To alleviate the problem of stragglers and speed up distributed SGD, we introduce an adaptive version of fastest-$k$ SGD with fixed step size that is communication-efficient. In fastest-$k$ SGD, as soon as the master updates the model $\mathbf{w}$ based on~\eqref{eeq6}, it broadcasts this model to all $n$ workers, and all workers start computing their partial gradients based on the new model, i.e., the ongoing computations of the slowest $n-k$ workers from the previous iteration are disregarded. Note that it is possible to obtain the partial computations of the slowest $n-k$ workers to improve the estimate of the full gradient; however, this improvement would come at the expense of a higher communication cost between the master and the workers which is problematic in our setting of interest. Furthermore, we study the case where the master distributes the data to the workers with no redundancy as a building step towards understanding adaptive implementations in the general case where the data is distributed with redundancy.

In the adaptive version of fastest-$k$ SGD which we propose, the master starts waiting for a small number of workers and increases this number as the algorithm evolves. The intuition behind varying $k$ is that at the beginning of the algorithm the master can boost SGD by waiting for a small number of workers, but it then has to increase this number gradually to enhance the accuracy of the model. Under standard assumptions on the loss function, and assuming independent and identical distributed random response times for the workers, we first give a theoretical bound on the error of fastest-$k$ SGD as a function of the wall-clock time. Based on this bound, we derive an adaptive policy that optimizes the error-runtime trade-off by increasing the value of $k$ at specific times which we explicitly determine in terms of the system parameters. We also devise an algorithm for adaptive fastest-$k$ SGD that is based on a statistical heuristic that works while being oblivious to the system parameters. We implement this algorithm and provide numerical simulations on synthetic data (linear regression) and real data (logistic regression on MNIST~\cite{deng2012mnist}). Our results show that our adaptive version of fastest-$k$ SGD outperforms non-adaptive implementations of fastest-$k$ SGD in terms of convergence rate, accuracy of the model, and also communication cost.

\section{Preliminaries}

We consider a random straggling model where the time spent by worker $i$ to finish the computation of its partial gradient (i.e., response time) is a random variable $X_{i}$, for $i=1,\ldots,n$. We assume that $X_{i}, i=1,\ldots, n$, are \emph{iid} and independent across iterations. Therefore, the time per iteration for fastest-$k$ SGD is given by the $k^{th}$ order statistic of the random variables $X_1,\ldots,X_n$, denoted by~$X_{(k)}$. In the previously described setting, the following bound on the error of fastest-$k$ SGD as a function of the number of iterations was shown in~\cite{Gauri,Bottou}.
\begin{proposition}[Error vs. iterations of fastest-$k$ SGD \cite{Gauri, Bottou}]
\label{prop1}
Under certain conditions (stated in the Appendix), the error of fastest-$k$ SGD after $j$ iterations with fixed step size satisfies
\begin{equation*}
\mathbb{E}\left[F(\mathbf{w}_j)-F^{\star} \right]\leq \frac{\eta L \sigma^2}{2cks} +(1-\eta c)^{j}  \bigg(F(\mathbf{w}_0)-F^{\star}-\frac{\eta L \sigma^2}{2cks} \bigg) \label{eeq10},
\end{equation*}
where $L$ and $c$ are the Lipschitz and the strong convexity parameters of the loss function respectively, $F^{\star}$ is the optimal value of the loss function, and $\sigma^2$ is the variance bound on the gradient estimate.
\end{proposition}

The notations used throughout the paper are summarized in Table~\ref{t}.

\begin{table}[h]
\centering
\setlength\extrarowheight{1.2pt}
 \begin{tabular}{C{1.5cm}|L{5.3cm}c C{1.5cm}|L{6.3cm}}
Variable & Description  & & Variable & Description \\ \cline{1-2} \cline{4-5}
$F$ & loss function & &  $\sigma^2$ & variance bound on the gradient estimate\\
$m$ & total number of data vectors & & $k$ & number of workers to wait for in fastest-$k$ SGD \\ 
$d$ & data dimension & & $X_i$ & random response time of worker $i$ \\ 
$n$ & number of workers & & $X_{(k)}$ & $k^{th}$ order statistic of $X_1,\ldots,X_n$ \\ 
$s$ & number of data vectors per worker  && $\mu_k$ & $\mathbb{E}[X_{(k)}]$: average of $X_{(k)}$   \\
$\eta$ & step size  &&  $\sigma_k^2$ & $Var[X_{(k)}]$: variance of $X_{(k)}$ \\ 
$L$ & Lipschitz constant  &&  $t$ & wall-clock time \\ 
$c$ & strong convexity parameter & &  $J(t)$ & number of iterations completed in time $t$  \\
\end{tabular}

\captionsetup{font=footnotesize}
\caption{\footnotesize{Summary of the notations used in the paper.}}
\vspace{-0.3cm}
\label{t}
\end{table}

\section{Theoretical Analysis}

In this section, we present our theoretical results. The proofs of these results are given in Section~\ref{proof}. In Lemma~\ref{lem1}, by applying techniques from renewal theory, we give a bound on the error of fastest-$k$ SGD as a function of the wall-clock time $t$ rather than the number of iterations. The bound holds with high probability for large $t$ and is based on Proposition~\ref{prop1}.
\begin{lemma} [Error vs. wall-clock time of fastest-$k$ SGD] 
\label{lem1}
Under the same assumptions as Proposition~\ref{prop1}, the error of fastest-$k$ SGD after wall-clock time $t$ with fixed step size satisfies
\begin{equation}
\label{eeq9}
\mathbb{E}\left[F(\mathbf{w}_t)-F^{\star} \right | J(t)]\leq  \frac{\eta L \sigma^2}{2cks}  + (1-\eta c)^{\frac{t}{\mu_k}(1-\epsilon)}  \left(F(\mathbf{w}_0)-F^{\star}-\frac{\eta L \sigma^2}{2cks} \right),
\end{equation}
with high probability \big($Pr \geq 1 - \frac{\sigma_k^2}{\epsilon^2}(\frac{2}{t\mu_k}+\frac{1}{t^2}) $\big) for large $t$, where $0<\epsilon \ll 1$ is a constant error term, $J(t)$ is the number of iterations completed in time $t$, and $\mu_k$ is the average of the $k^{th}$ order statistic $X_{(k)}$.
\end{lemma}

Notice that the first term in~\eqref{eeq9} is constant (independent of $t$), whereas the second term decreases exponentially in~$t$ ($\eta c<1$ from~\cite{Bottou}). In fact, it is well known that the SGD with constant step size goes first through a transient phase where the error decreases exponentially fast, and then enters a stationary phase where the error oscillates around a constant term~\cite{Murata}. From~\eqref{eeq9}, it is easy to see that the rate of the exponential decrease in the transient phase is governed by the value of $1/\mu_k$. Since $\mu_k$ is an increasing function of $k$, the exponential decrease is fastest for $k=1$ and slowest for $k=n$. Whereas the stationary phase error which is upper bounded by $\eta L \sigma^2/2cks$, is highest for $k=1$ and lowest for $k=n$. This creates a trade-off between the rate of decrease of the error in the transient phase, and the error floor achieved in the stationary phase. Ultimately, we would like to first have a fast decrease through the transient phase, and then have a low error in the stationary phase. To this end, we look for an adaptive policy for varying $k$ that starts with $k=1$ and then switches to higher values of $k$ at specific times in order to optimize the error-runtime trade-off. Such an adaptive policy guarantees that the error is minimized at every instant of the wall-clock time $t$.

Since the bound in~\eqref{eeq9} holds with high probability for large $t$, we explicitly derive the switching times for adaptive fastest-$k$ SGD that optimize this bound. Note that for the sake of simplicity, we drop the constant error term $\epsilon$ in our analysis.

\begin{theorem}[Bound-optimal policy for adaptive fastest-$k$ SGD]
\label{thm2}
The bound-optimal times $t_k, k=1, \ldots, n-1$, at which the master should switch from waiting for the fastest $k$ workers to waiting for the fastest $k+1$ workers are given by
\begin{equation*}
t_k  = t_{k-1}+\frac{\mu_k}{-\ln (1-\eta c)} \times  \bigg[ \ln \left(\mu_{k+1}-\mu_k \right)   -  \ln \left(\eta L \sigma^2\mu_k \right) +\ln \left(2ck(k+1)s(F(\mathbf{w}_{t_{k-1}})-F^{\star})-\eta L (k+1) \sigma^2) \right)\bigg] \label{eqq11},
\end{equation*} 
where $t_0=0$.
\end{theorem}

\begin{example}[Theoretical analysis on fastest-$k$ SGD with {\em iid} exponential response times] 
\label{ex2}
Suppose $X_i\sim \exp(\lambda)$, $i=1,\ldots,n$, is the random time taken by worker~$i$ to compute its partial gradient, where $X_i$'s are iid. The average time spent by the master per iteration in fastest-$k$ SGD is the average of the $k^{th}$ order statistic of the exponential random variables $X_1,\ldots, X_n$, given by
\begin{equation}
\label{eq3}
\mu_k=\frac{1}{\lambda}\sum_{i=n-k+1}^{n}\frac{1}{i}.
\end{equation}
 Let $n=5, \lambda=1/\mu=1/5, \eta=0.001, \sigma^2=10, F(\mathbf{w}_0)-F^{\star}=100, L=2, c=1, s=10$. We evaluate the bound in Lemma~\ref{lem1} for multiple fixed values of $k$ (non-adaptive) and compare it to adaptive fastest-$k$ SGD if we apply the switching times in Theorem~\ref{thm2}. Let $k_{fixed}^{\star}(t)$ and $k_{adaptive}^{\star}(t)$ be the bound-optimal choices of the number of workers to wait for as a function of the wall-clock time for non-adaptive and adaptive fastest-$k$ SGD, respectively. The results are shown in Fig.~\ref{fig1}~and~\ref{fig2}.

\begin{figure}[h]
\centering
\includegraphics[width=0.5\textwidth]{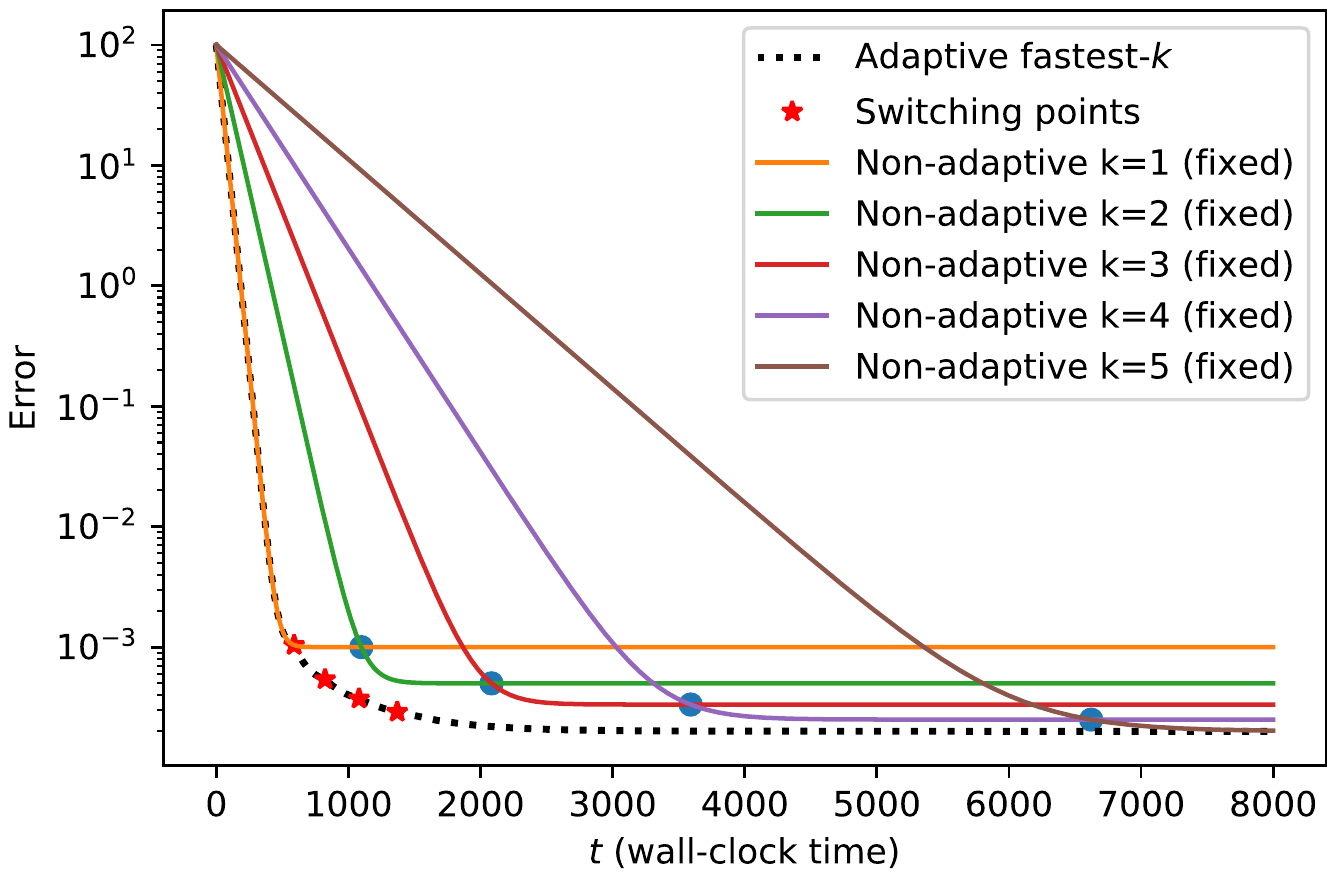}
\caption{The upper bound on the error given by~\eqref{eeq9} as a function of time, evaluated for $k=1,2,3,4,5$.}
\label{fig1}
\end{figure}

\begin{figure}[h!]
\vspace{-0.5cm}
\centering
\begin{subfigure}[h!]{0.49\textwidth}
\centering
\includegraphics[width=0.97\textwidth]{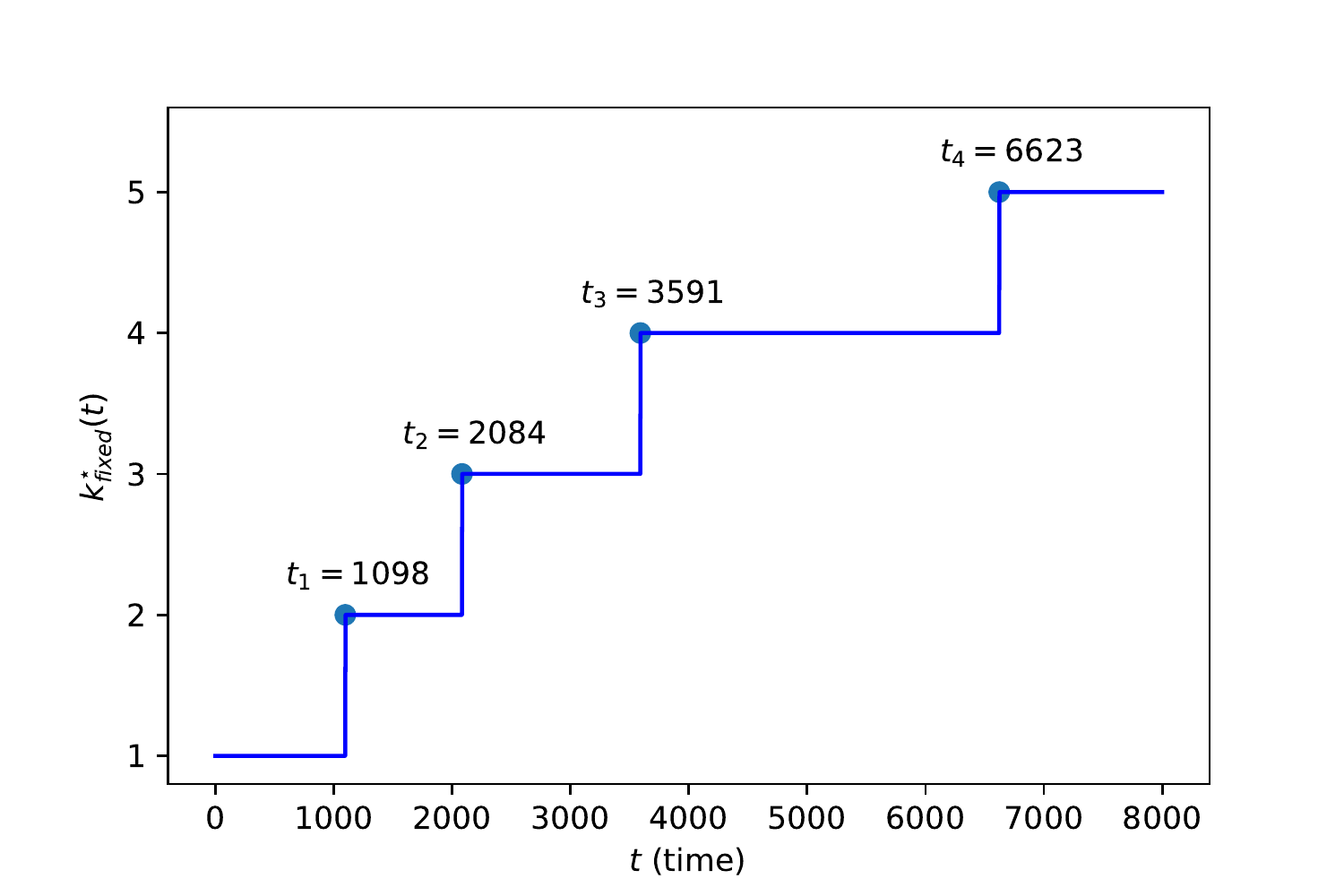}
\caption{Non-adaptive fastest-$k$ SGD.}
\label{fig2a}
\end{subfigure} 
\hspace{0.1cm}
\begin{subfigure}[h!]{0.49\textwidth}
\centering
\includegraphics[width=0.97\textwidth]{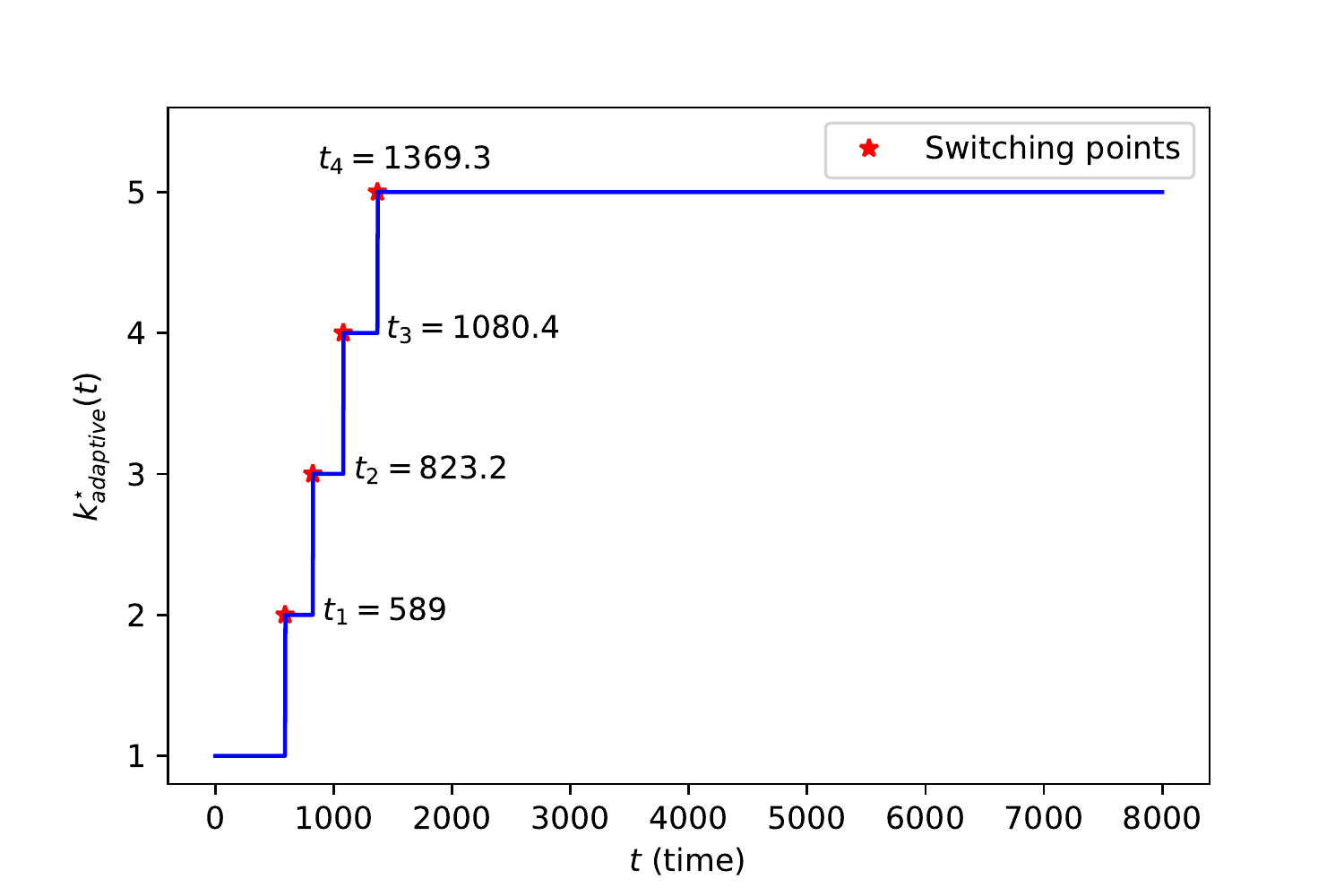}
\caption{Adaptive fastest-$k$ SGD.}
\label{fig2b}
\end{subfigure}
\caption{The bound-optimal choices of the number of workers to wait for $k_{fixed}^{\star}(t)$ and $k_{adaptive}^{\star}(t)$ for non-adaptive and adaptive fastest-$k$ SGD, respectively.}
\label{fig2}
\end{figure}

The plots in Fig.~\ref{fig1} and Fig.~\ref{fig2a} show that for the time interval $[0,1098)$, the bound-optimal choice of fixed (non-adaptive) $k$ that minimizes the error is $k=1$. Namely, if one wants to run non-adaptive fastest-$k$ SGD for less than $1098$ time units, then the best strategy is to wait for the fastest single worker before updating the model. It is easy to see from Fig.~\ref{fig1} that any other choice of fixed $k>1$, leads to a higher error in the time interval $[0,1098)$. For a runtime greater than $6623$ time units, the bound-optimal choice for non-adaptive fastest-$k$ SGD is to wait for the responses of all workers. Whereas for adaptive fastest-$k$ SGD, notice from Fig.~\ref{fig1} and Fig.~\ref{fig2b}, that for the time interval $[0,589)$, the adaptive policy assigns $k=1$ since it gives the fastest error decrease in the beginning. Then, as the error approaches the stationary phase, the policy increases $k$ to $k=2$. This allows the error to decrease below the error floor for $k=1$. The procedure continues until $k$ attains its maximum value of $k=n=5$. In conclusion, the results demonstrate that the adaptive version enables achieving lower error values in less time.

\end{example}

This analysis shows the potential of adaptive strategies in optimizing the error-runtime trade-off. It also suggests that the value of $k$ should be gradually increased throughout the runtime of fastest-$k$ SGD in order to optimize this trade-off. Although Theorem~\ref{thm2} provides useful insights about how to adapt $k$ over time, applying the exact switching times derived in Theorem~\ref{thm2} may not be effective in practice due to the following reasons: \begin{enumerate*}[label={(\roman*)}] \item the policy optimizes an upper bound on the error (Lemma~\ref{lem1}) which is probabilistic and may be loose; \item the policy requires the knowledge of several system parameters including the optimal value of the loss function $F^{\star}$ which is typically unknown.\end{enumerate*} Nevertheless, we use the insights provided by the theoretical analysis to design a practical algorithm for adaptive fastest-$k$ SGD. This algorithm is based on a statistical heuristic and is oblivious to the system parameters as we explain in Section~\ref{sec5}.

\section{Adaptive Fastest-$k$ SGD Algorithm}
\label{sec5}

In this section we present an algorithm for adaptive fastest-$k$ SGD that is realizable in practice. As previously mentioned, SGD with fixed step size goes first through a transient phase where the error decreases exponentially fast, and then enters a stationary phase where the error oscillates around a constant term. Initially, the exponential decrease is fastest for $k=1$. Then, as the stationary phase approaches, the error decrease becomes slower and slower until a point where the error starts oscillating around a constant term and does not decrease any further. At this point, increasing $k$ allows the error to decrease further because the master would receive more partial gradients and hence would obtain a better estimate of the full gradient. The goal of the adaptive policy is to detect this phase transition in order to increase $k$ and keep the error decreasing.

The adaptive policy we present in this section detects this phase transition by employing a statistical test based on a modified version of Pflug's procedure for stochastic approximation~\cite{Pflug}. The main component of our policy is to monitor the signs of the products of consecutive gradients computed by the master based on~\eqref{eeq6}. The underlying idea is that in the transient phase, due to the exponential decrease of the error, the gradients are likely to point in the same direction, hence their inner product is positive. Our policy consists of utilizing a counter that counts the difference between the number of times the product of consecutive gradients is negative (i.e., ${\ghat_j}^T \ghat_{j-1} < 0$) and the number of times this product is positive, throughout the iterations of the algorithm. 

In the beginning of the algorithm, we expect the value of the counter to be negative and decrease because of the exponential decrease in the error. Then, as the error starts moving towards the stationary phase, negative gradient products will start accumulating until the value of the counter becomes larger than a certain positive threshold. At this point, we declare a phase transition and increase $k$. The complete algorithm is given in Algorithm~\ref{algo_disjdecomp}.

\begin{algorithm}[h]
\small
\SetKwData{Left}{left}
\SetKwData{This}{this}
\SetKwData{Money}{money}
\SetKwData{MaxIter}{maxIter}
\SetKwData{Step}{step}
\SetKwData{CountI}{countIter}
\SetKwData{CountN}{countNegative}
\SetKwData{Thresh}{thresh}
\SetKwData{Burnin}{burnin}
\SetKwData{Up}{up}
\SetKwFunction{Budget}{budget}
\SetKwFunction{Union}{Union}
\SetKwFunction{FindCompress}{FindCompress}
\SetKwInOut{Input}{input}
\SetKwInOut{Output}{output}
\Input{starting point $\mathbf{w}_0$, data $\{X,\mathbf{y}\}$, number of workers $n$, step size $\eta$, maximum number of iterations~$J$, adaptation parameters \Step , \Thresh, \Burnin }
\Output{weight vector $\mathbf{w}_J$}
$j\gets 1$ \\
$k \gets 1$ \\
$\CountN \gets 0$ \\
$\CountI \gets 1$ \\
Distribute $X$ to the $n$ workers \\
\While{$j\leq J$}{
Send $\mathbf{w}_{j-1}$ to all workers \\
Collect the responses of the fastest $k$ workers \\
$\mathbf{w}_{j}\gets \mathbf{w}_{j-1}-\eta \ghat_{j-1}$ \\
\eIf{${\ghat_j}^T \ghat_{j-1}  < 0$ } { 
$\CountN \gets \CountN + 1$} { $\CountN \gets \CountN - 1$}

\If{$\CountN>\Thresh$ {\bf and} $\CountI>\Burnin$ {\bf and} $k\leq n-\Step$} {
$k\gets k+\Step$ \\
$\CountN\gets 0$ \\
$\CountI\gets 0$ 
}
$\CountI \gets \CountI + 1$ \\
$j\gets j+1$ 
}
\Return $\mathbf{w}_J$ 
\label{alg1}
\caption{Adaptive Fastest-$k$ SGD}\label{algo_disjdecomp}
\end{algorithm}

\section{Simulations}
\label{sec6}
\subsection{Linear regression on synthetic data}
\label{synthetic}
\emph{Experimental setup.} We simulated the performance of the fastest-$k$ SGD (Algorithm~\ref{algo_disjdecomp}) described earlier for $n$ workers on synthetic data $X$. We generated $X$ as follows: \begin{enumerate*}[label={(\roman*)}] \item we pick each row vector $\mathbf{x}_{\ell}$, $\ell=1,\dots,m,$ independently and uniformly at random from $\{1,2,\dots,10\}^{d}$;  \item we pick a random vector $\bar{\mathbf{w}}$ with entries being integers chosen uniformly at random from $\{1,\dots,100\}$; and \item we generate $\mathbf{y}_{\ell} \sim \mathcal{N}(\ip{\mathbf{x}_{\ell}}{\bar{\mathbf{w}}},1)$ for all $\ell=1,\dots,m$.\end{enumerate*} 

We run linear regression using the $\ell_2$ loss function, i.e., $F(\mathbf{w}) = \displaystyle \sum_{\ell=1}^{m} \dfrac{1}{2} \norm{\ip{\mathbf{x}_{\ell}}{\mathbf{w}} - \mathbf{y}_{\ell}}_2^2$. At each iteration, we generate $n$ independent exponential random variables with rate $\lambda = 1$.

\begin{figure}[h!]
\centering
\includegraphics[width=0.55\textwidth]{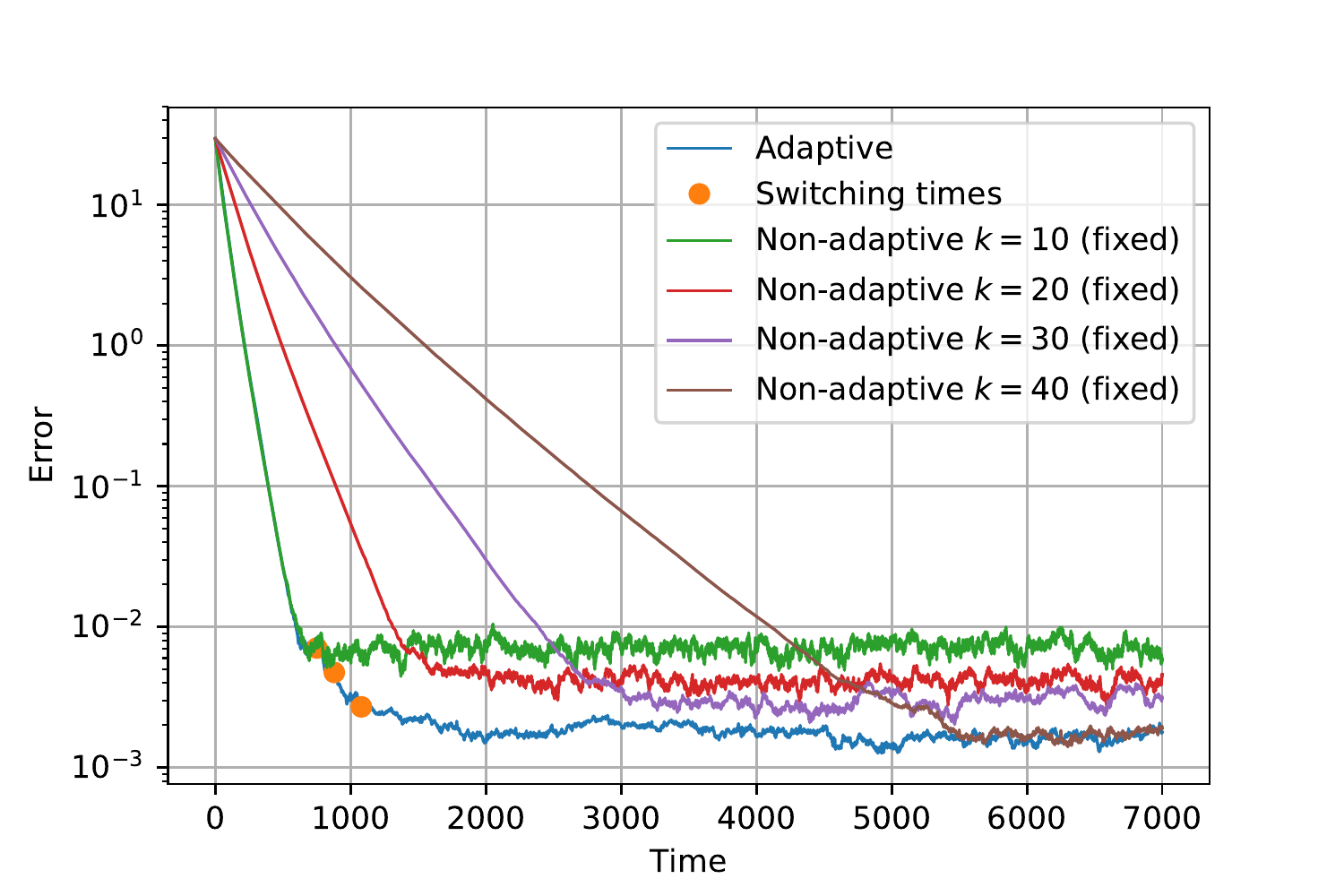}
\caption{\small Error as a function of the wall-clock time for non-adaptive fastest-$k$ SGD with fixed $k=10,20,30,40$; and adaptive fastest-$k$ SGD (Algorithm~1). The experimental setup is the following: $d=100$, $m=2000$, $n=50$, $\eta=0.0005$. The adaptation parameters chosen here are $step=10$, $thresh=10$, and $burnin=0.1\times $(number of data points)=200. We start the adaptive fastest-$k$ SGD with $k=10$ and increase $k$ by $10$ until reaching $k=40$.}
\label{fig3}
\end{figure}

\emph{Observations.} Figure~\ref{fig3} compares the performance of the adaptive fastest-$k$ SGD its non-adaptive counterpart for $n=50$ workers and different values of $k$. In the adaptive version we start with $k=10$ and then increase $k$ by $10$ until reaching $k=40$, where the switching times are given by Algorithm~\ref{algo_disjdecomp}. Whereas for the non-adaptive version, $k$ is fixed throughout the runtime of the algorithm. We run fastest-$k$ SGD for $k\in\{10,20,30,40\}$. The comparison shows that the adaptive version is able to achieve a better error-runtime trade-off than the non-adaptive ones. Namely, notice that the adaptive fastest-$k$ SGD reaches its lowest error at approximately $t=2000$, whereas among the non-adaptive versions the same error is only reached for $k=40$ at approximately $t=6000$. These results confirm our intuition and previous theoretical results.

\begin{figure}[h!]
\vspace{-0.5cm}
\centering
\begin{subfigure}[h!]{0.49\textwidth}
\centering
\includegraphics[width=0.97\textwidth]{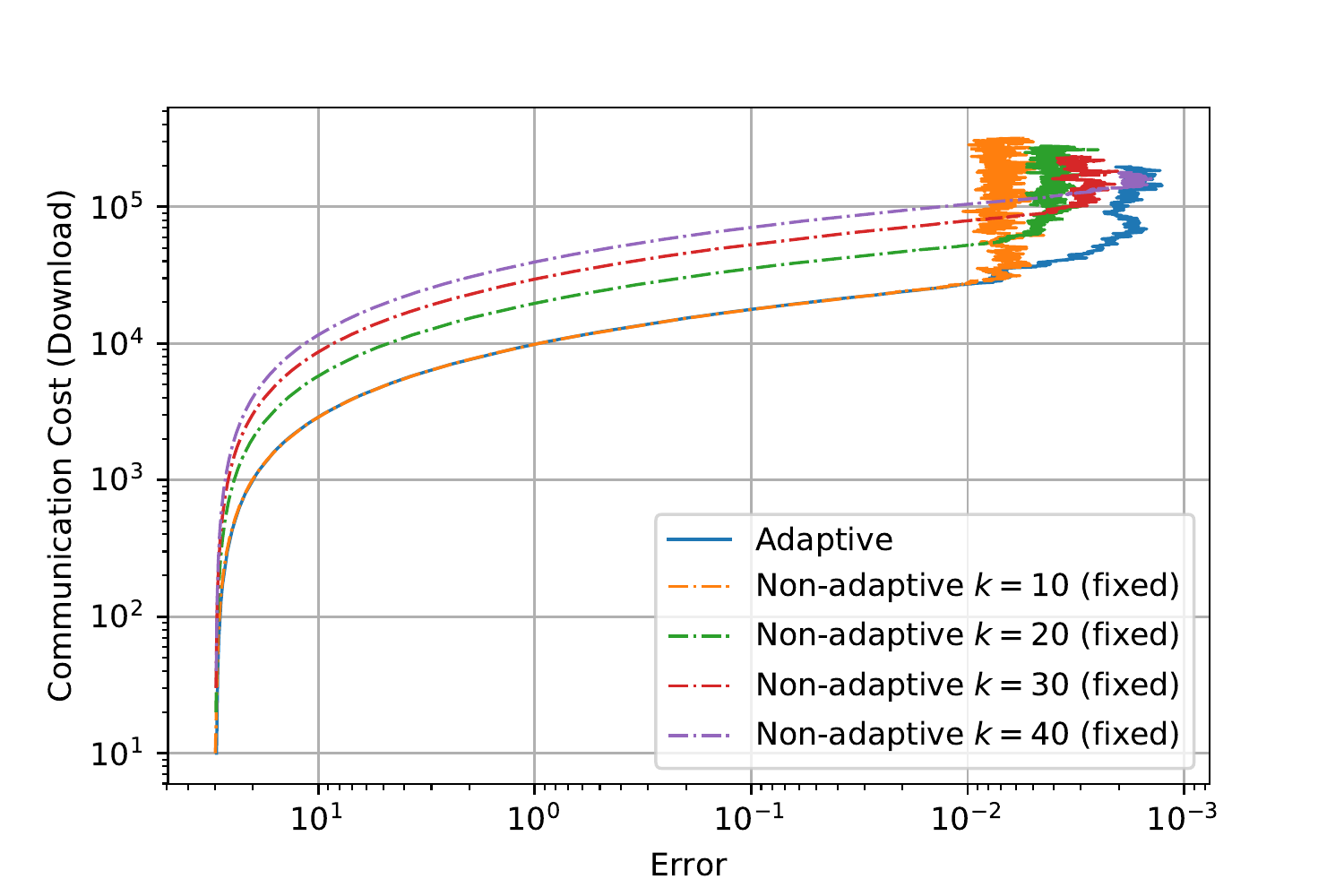}
\caption{Download}
\label{fig4a}
\end{subfigure} 
\hspace{0.1cm}
\begin{subfigure}[h!]{0.49\textwidth}
\centering
\includegraphics[width=0.97\textwidth]{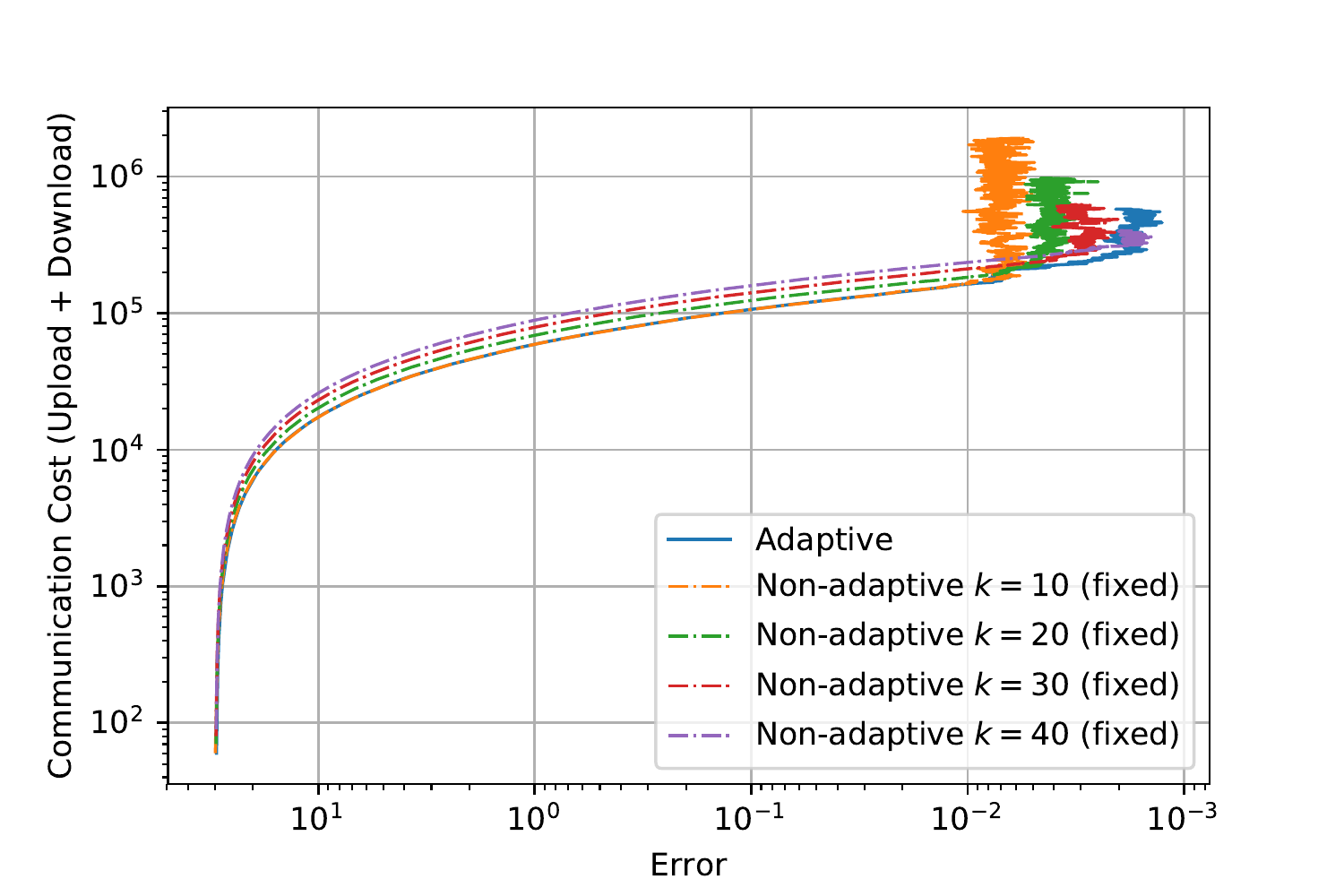}
\caption{Upload + Download}
\label{fig4b}
\end{subfigure}
\caption{Communication cost as a function of the error for the same setting described in Figure~\ref{fig3}.}
\label{com1}
\end{figure}

Furthermore, in Figure~\ref{com1} we analyze the communication cost incurred by the learning process to reach a certain error, equivalently to reach a certain desired accuracy. The communication cost is measured by the number of vectors in $\mathbb{R}^d$ that are communicated in order to reach a given error value, where each vector in $\mathbb{R}^d$ represents one unit of communication. The download communication cost is measured in terms of the number of partial gradients that the master downloads from the workers over a point-to-point link. Whereas, the upload communication cost is measured in terms of the number of model vectors $\mathbf{w}$ that are broadcasted by the master to the workers. Since at every iteration of the algorithm all workers receive the updated model, the upload cost is equal to $n$ units of communication per iteration. Note that in certain settings, the cost of broadcasting the same model vector $\mathbf{w}$ to all workers may be cheap compared to downloading different partial gradients from the workers. For this reason, we plot the download communication cost separately in Figure~\ref{fig4a} and plot the total communication cost (upload + download) in Figure~\ref{fig4b}. The results show that the adaptive fastest-$k$ SGD requires less communication than the non-adaptive versions for reaching any given error value.

\newcommand{\target}{\ensuremath{u}}
\subsection{Logistic regression on MNIST}
\emph{Setup.} We simulated the performance of the fastest-$k$ SGD (Algorithm~\ref{algo_disjdecomp}) for $n=20$ workers on the MNIST handwritten dataset~\cite{deng2012mnist}. For this experiment, we used only $m=80$ images as training data. The MNIST dataset consists of handwritten digits. Every digit is represented by a $28\times 28$ pixels image. The images are transformed to vectors of length $784$ each. Each vector $\mathbf{x}_\ell$ has a label $y_\ell \in \{0,\dots,9\}$ reflecting which digit is represented by this image.

We run logistic regression. The goal is to find ten model vectors $\mathbf{w}_\target$ of length $784$ each, $\target\in \{0,\dots,9\}$, and a bias vector $\mathbf{b}$ of length $10$ such that for every $\ell$ the function $\sigma(\left\langle\mathbf{x}_\ell,\mathbf{w}_\target \right\rangle, b_\target) \triangleq \frac{1}{1+\exp(b_\target+\left\langle\mathbf{x}_\ell,\mathbf{w}_\target \right\rangle)}$ is maximum for $\target = y_\ell$. The function $\sigma(\left\langle\mathbf{x}_\ell,\mathbf{w}_\target \right\rangle, b_\target)$ represents the probability of image $\ell$ being digit $\target$ and should be maximum for the true label $y_\ell$. For every label $\target \in \{0,\dots,9\}$, the vector $\mathbf{y}$ is encoded into a new vector $\mathbf{y}^\target$. The $\ell$th entry ${y}^\target_\ell$ of $\mathbf{y}^\target$ is equal to one if $y_\ell = \target$ and zero otherwise. The loss function is 

\begin{equation}\label{eq:avg_ll}
    F(\mathbf{w}, \mathbf{b}) = -\frac{1}{10}\sum_{\target = 0}^9 \sum_{\ell = 1}^{m} y^\target_\ell \log (\sigma (\left\langle\mathbf{x}_\ell,\mathbf{w}_\target \right\rangle, b_\target) ) + (1-y^\target_\ell) \log(\sigma (\left\langle\mathbf{x}_\ell,\mathbf{w}_\target \right\rangle, b_\target) ).
\end{equation}

We run a regularized logistic regression with the regularizer being the $\ell_2$ norm of $\mathbf{w}_\target$. In other words, the gradient computed at every iteration for every label $\target$ is that of $F(\mathbf{w}_\target,b_\target)+ \frac{\lambda}{2}\norm{\mathbf{w}_\target}_2^2$. At every iteration, we generate $n$ independent exponential random variables with rate $\lambda = 1/50$.

\begin{figure}[h!]
\centering
\includegraphics[width=0.5\textwidth]{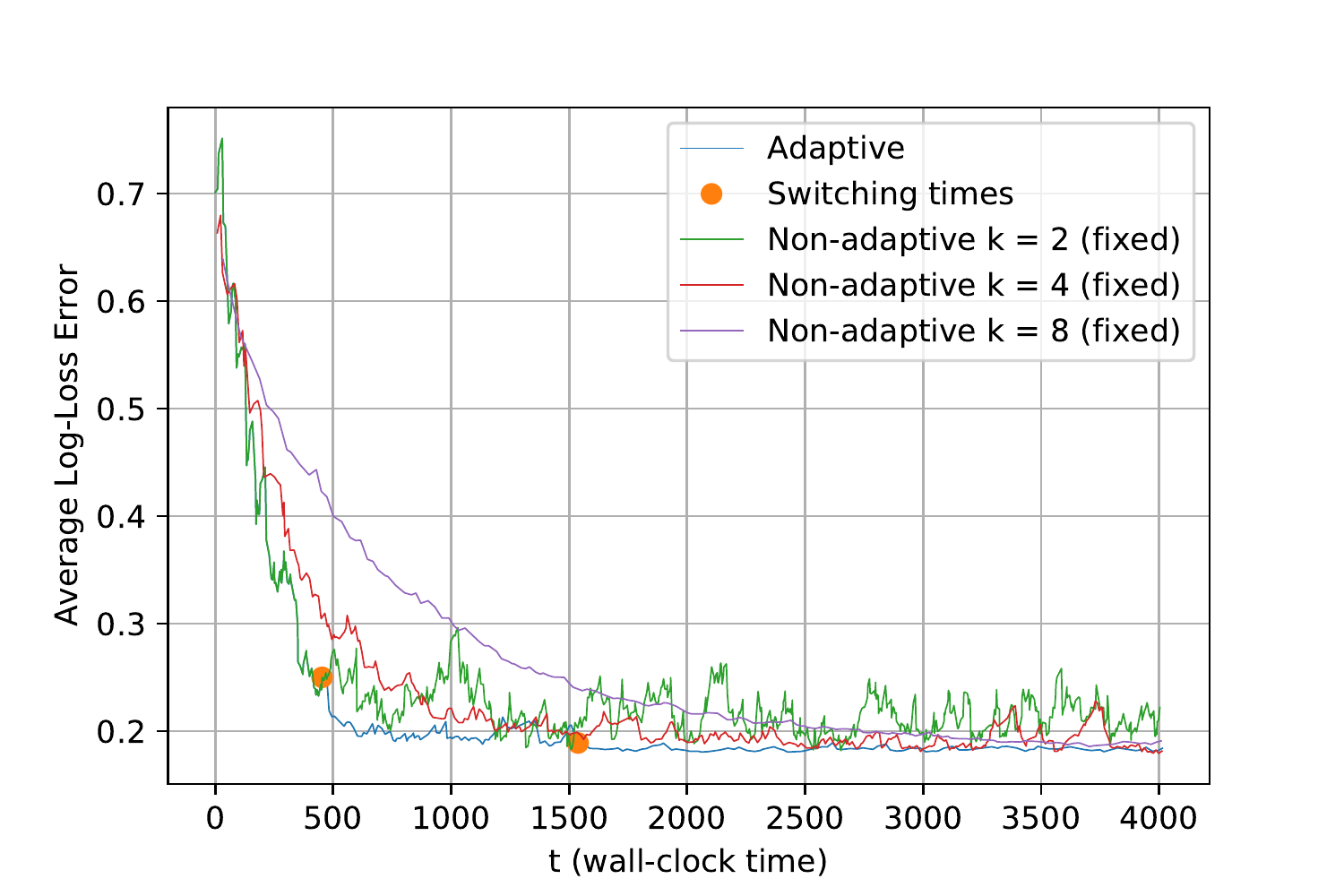}
\caption{\small Average log-loss error~\eqref{eq:avg_ll} as a function of the wall-clock time for non-adaptive fastest-$k$ SGD with fixed $k=2,4,8$; and adaptive fastest-$k$ SGD (Algorithm~1). The experimental setup is the following: $d=784$, $m=60$, $n=20$, $\eta=0.05$ and $\lambda = 0.01$. The adaptation parameters chosen here are, $thresh=20$, and $burnin=30$. We start the adaptive fastest-$k$ SGD with $k=2$ and double $k$ until reaching $k=8$.}
\label{fig:MNIST-error}
\end{figure}

\emph{Observations.} Figure~\ref{fig:MNIST-error} compares the performance of the adaptive fastest-$k$ SGD to its non-adaptive counterpart for $n=20$ workers and different values of $k$. In the adaptive version we start with $k=2$ and then double $k$ until reaching $k=8$, where the switching times are given by Algorithm~\ref{algo_disjdecomp}. Whereas for the non-adaptive version, $k$ is fixed throughout the runtime of the algorithm. We run fastest-$k$ SGD for $k\in\{2,4,8\}$. Similar to the case of synthetic data in Section~\ref{synthetic}, the comparison shows that the adaptive version is able to achieve a better error-runtime trade-off than the non-adaptive ones. Namely, that the adaptive fastest-$k$ SGD reaches its lowest error at approximately $t=1700$, whereas the non-adaptive versions either have higher errors or take much longer time to reach the same error. 

\begin{figure}[h!]
\vspace{-0.5cm}
\centering
\begin{subfigure}[h!]{0.49\textwidth}
\centering
\includegraphics[width=0.97\textwidth]{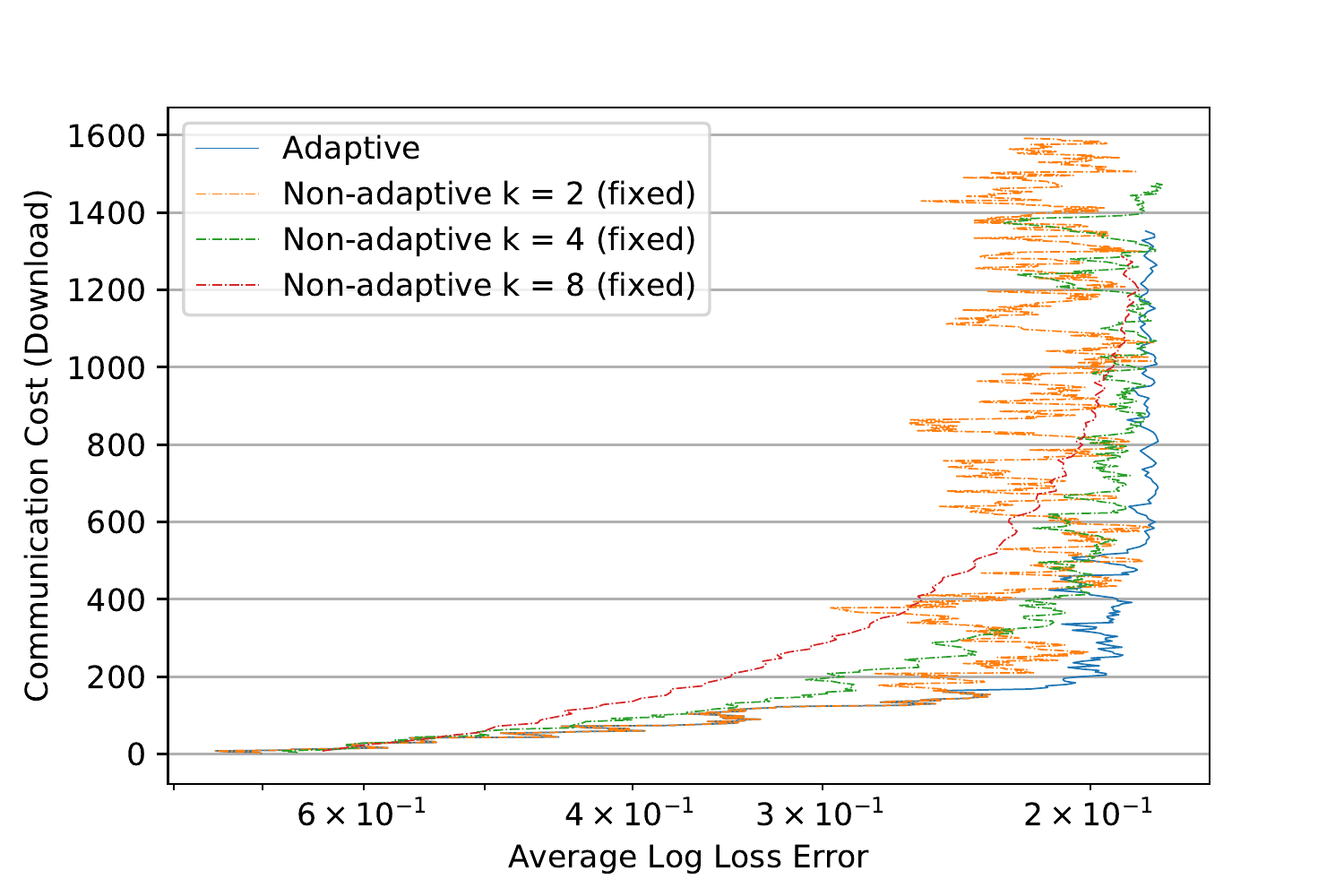}
\caption{Download}
\label{fig5a}
\end{subfigure} 
\hspace{0.1cm}
\begin{subfigure}[h!]{0.49\textwidth}
\centering
\includegraphics[width=0.97\textwidth]{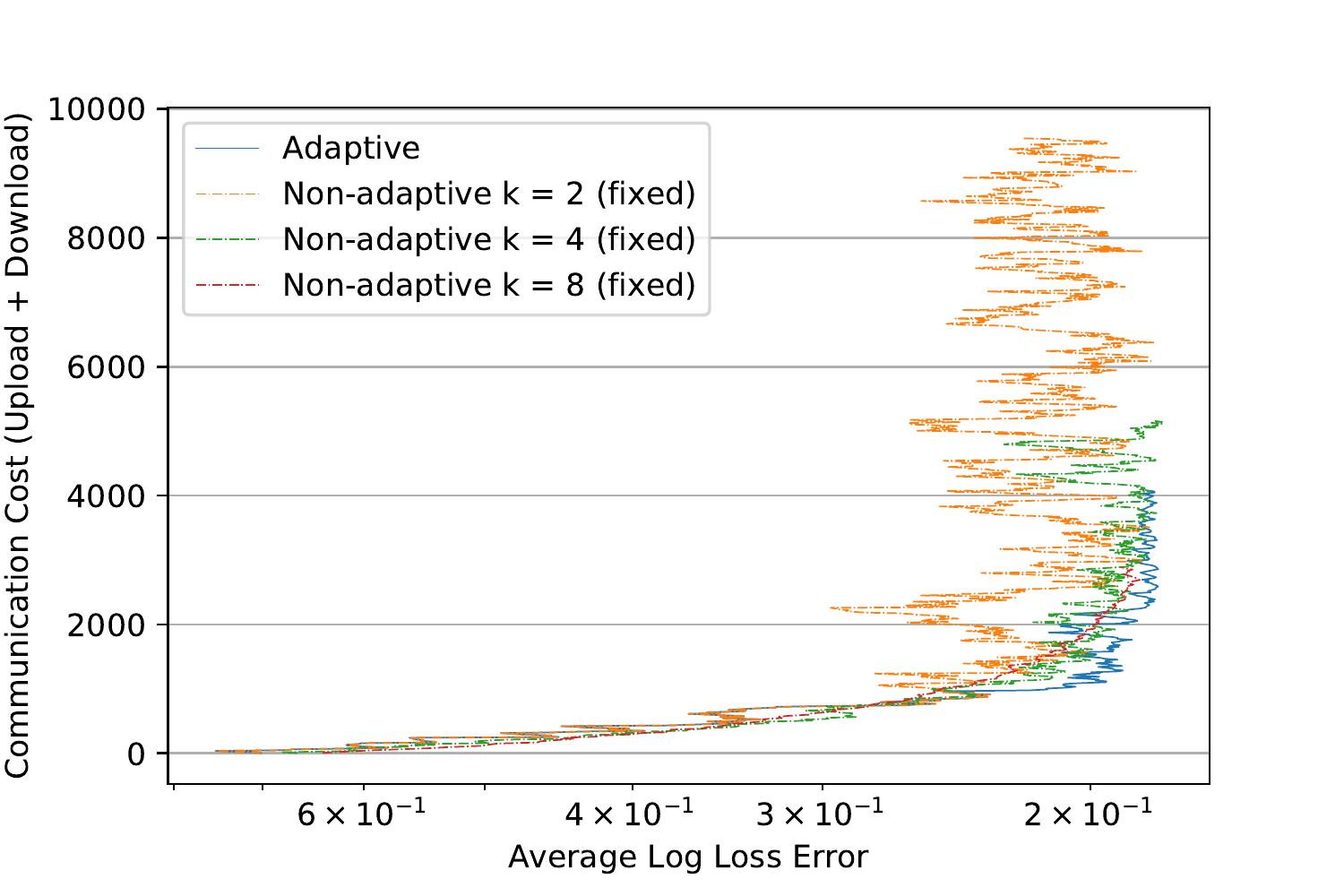}
\caption{Upload + Download}
\label{fig5b}
\end{subfigure}
\caption{Communication cost as a function of the average log-loss \eqref{eq:avg_ll} for the same setting described in Figure~\ref{fig:MNIST-error}.}
\label{com2}
\end{figure}

Figure~\ref{com2} compares the communication cost of adaptive fastest-$k$ SGD to the non-adaptive version as a function of the reached average log-loss~\eqref{eq:avg_ll}. Similar to the case of synthetic data in Section~\ref{synthetic}, for a fixed desired error target, the adaptive fastest-$k$ SGD requires the smallest download communication cost as shown in Figure~\ref{fig5a}. As for the download + upload communication cost shown in~\ref{fig5b}, this cost for the adaptive fastest-$k$ SGD is slightly higher than that of the non-adaptive counterparts for high error values. However, for low error values (higher accuracy) that are close to $2\times 10^{-1}$, the adaptive version has the lowest communication cost and is the fastest algorithm.

\section{Proofs}
\label{proof}
\subsection{Proof of Lemma~\ref{lem1}}
Proposition~\ref{prop1} gives an upper bound on the error of fastest-$k$ SGD as a function of the number of iterations $j$, given the conditions in the Appendix. Let $J(t)$ be the arrival process representing the number of iterations completed in time~$t$. The upper bound as a function of time is given by 
\begin{equation}
\label{eeeq9}
\mathbb{E}\left[F(\mathbf{w}_t)-F^{\star}| J(t) \right]\leq  \frac{\eta L \sigma^2}{2cks}+ (1-\eta c)^{J(t)}  \left(F(\mathbf{w}_0)-F^{\star}-\frac{\eta L \sigma^2}{2cks} \right).
\end{equation}
The time between any two consecutive iterations in fastest-$k$ SGD is given by $X_{(k)}$, where $X_{(k)}$ is the $k^{th}$ order statistic of the random variables $X_1, \ldots, X_n$. Therefore, $J(t)$ has {\em iid} inter-arrival times. Let $(T_z \in \R_+: z \in \N)$ be the {\em iid} sequence of inter-arrival times, where $\mathbb{E}[T_z]=\mu_k$ and $Var[T_z]=\sigma_k^2$.  We define $S_j = \sum_{z=1}^j T_z$, such that $S_0 = 0$ and $S_j$ is the $j$th renewal instant for \emph{iid} sequence $T = (T_z: z \in \N)$ with mean $\mu_k$, variance $\sigma_k^2$, and the moment generating function $M_T(\alpha) = e^{\Phi_T(\alpha)}$.   
Then, we can write the following concentration inequalities for $S_j$. 
First, from the Chebyshev inequality we have 
\begin{equation}
P\set{\abs{\frac{S_j - j\mu_k}{\sqrt{n}\sigma_k}} > t} \le \frac{1}{t^2}. 
\end{equation}
Corresponding to the renewal process, we obtain the counting process 
\begin{equation}
J(t) = \sum_{z \in \N}\indicator{T_z \le t}.
\end{equation}
Defining $x_l \triangleq y_l\frac{\sigma_k}{\mu_k}\sqrt{\frac{t}{\mu_k}}$ and $x_u \triangleq y_u\frac{\sigma_k}{\mu_k}\sqrt{\frac{t}{\mu_k}}$, 
we can bound from above the union of the events $\set{J(t) > \frac{t}{\mu_k} + x_u}$ and $\set{J(t) < \frac{t}{\mu_k} -x_l}$. 
Since $J(t)$ is integer valued, taking the floor $n_u = \lfloor \frac{t}{\mu_k} + x_u \rfloor$ and the ceil $n_l = \lceil \frac{t}{\mu_k} - x_l \rceil$, we can write using the inverse relationship between the counting process $(J(t): t \ge 0)$ and the renewal process $(S_n: n \in \N)$ as
\begin{align}
P\set{J(t) > \frac{t}{\mu_k} + x_u} &= \set{S_{n_u} - \mu_k n_u \le t - \mu_k n_u} = P\set{\frac{S_{n_u} - \mu_k n_u}{\sqrt{n_u}\sigma_k} \le -\frac{\mu_k x_u}{\sqrt{n_u}\sigma_k}} \le \frac{\sigma_k^2n_u}{\mu_k^2x_u^2}, \\
P\set{J(t) < \frac{t}{\mu_k} - x_l} &= \set{S_{n_l} - \mu_k n_l \ge t - \mu_k n_l} = P\set{\frac{S_{n_l} - \mu_k n_l}{\sqrt{n_l}\sigma_k} \ge \frac{\mu_k x_l}{\sqrt{n_l}\sigma_k}} \le \frac{\sigma_k^2n_l}{\mu_k^2x_l^2}.
\end{align}
Substituting the value of $x_u,x_l$ in terms of $y_u, y_l$, we get 
\begin{equation}
P\set{-y_l  \le \frac{J(t)- \frac{t}{\mu_k}}{\sigma_k \sqrt{\frac{t}{\mu_k^3}}} \le y_u} \ge 1- \left(\frac{1}{y_l^2} + \frac{1}{y_u^2} + \frac{\mu_k}{ty_l^2}\right) - \frac{\sigma_k}{\sqrt{t \mu_k}}\left(\frac{1}{y_u} - \frac{1}{y_l}\right). 
\end{equation}
Let $\alpha \triangleq -\ln(1-\eta c) > 0$,
We can write the following bound for $(1-\eta c)^{J(t)} = e^{-\alpha J(t)}$. 
\begin{equation}
\label{eq11}
P\set{e^{-\alpha\frac{t}{\mu_k}(1 + \frac{y\sigma_k}{\sqrt{t\mu_k}})} \le e^{-\alpha J(t)} \le e^{-\alpha\frac{t}{\mu_k}(1 - \frac{y\sigma_k}{\sqrt{t\mu_k}})}} \ge 1 - \left(\frac{\frac{\mu_k}{t}+2}{y^2}\right).
\end{equation}
By substituting $y=\epsilon \frac{\sqrt{t \mu_k}}{\sigma_k}$ in~\eqref{eq11}, where $0<\epsilon \ll 1$ is a constant, we obtain 
\begin{equation}
\label{eq12}
Pr\bigg\{ (1-\eta c)^{J(t)} \le (1-\eta c)^{\frac{t}{\mu_k}(1-\epsilon)} \bigg\} \geq 1 - \frac{\sigma_k^2}{\epsilon^2}\left(\frac{2}{t\mu_k}+\frac{1}{t^2} \right).
\end{equation}
From~\eqref{eeeq9}~and~\eqref{eq12}, we can see that that the bound \eqref{eeq9} in Lemma~\eqref{lem1} holds with high probability for large $t$.
\subsection{Proof of Theorem~\ref{thm2}}
To minimize the error achieved within any time interval, we maximize the convergence rate at every time $t$, based on the result in Lemma~\ref{lem1} (the error term $\epsilon$ is ommitted). 
Substituting $\alpha \triangleq -\ln(1-\eta c)>0$ in~\eqref{eeq9}, we define
\begin{equation}
\label{eeq13}
e(t) \triangleq  \frac{\eta L \sigma^2}{2cks} + e^{ \frac{-\alpha t}{\mu_k}}  \left(F(\mathbf{w}_0)-F^{\star}-\frac{\eta L \sigma^2}{2cks} \right),
\end{equation}
where $e(0)=F(\mathbf{w}_0)-F^{\star}$. To evaluate the rate of convergence, we analyze the magnitude of the first order derivative of $e(t)$ with respect to $t$ given by
\begin{align}
\label{eeq14}
\left \lvert \frac{d e(t)}{d t} \right \rvert = 
\frac{\alpha}{\mu_k} e^{ \frac{-\alpha t}{\mu_k}} \left(e(0)-\frac{\eta L \sigma^2}{2cks} \right).
\end{align}
For a fixed $t'\leq t$, we can write the following equivalent definitions of~\eqref{eeq13} and~\eqref{eeq14}.
\begin{equation}
e(t) = \frac{\eta L \sigma^2}{2cks}+e^{ \frac{-\alpha (t-t')}{\mu_k} } \left(e(t')-\frac{\eta L \sigma^2}{2cks} \right). 
\end{equation}
\begin{equation}
\label{eeq17}
\left \lvert \frac{d e(t)}{d t} \right \rvert = \frac{\alpha}{\mu_k} e^{ \frac{-\alpha(t-t')}{\mu_k}} \left(e(t')-\frac{\eta L \sigma^2}{2cks} \right).
\end{equation}
The value of $k$ in the expressions above is considered to be fixed. We are interested in an adaptive policy for choosing $k$ that maximizes this magnitude at every time $t$, and hence we look for a scheduling function $k(t)$ that varies $k$ with time. We observe that there is trade-off between the two terms of $e(t)$ in~\eqref{eeq13}. Waiting for larger number of workers $k$ increases the mean iteration time $\mu_k$, whereas it decreases the stationary phase error (i.e., error floor)~$\eta L\sigma^2/2cks$. Assuming that the initial error $e(0)$ is much larger than the stationary phase error, it is easy to see that the rate of decrease in error at $t=0$ is highest for $k=1$. Hence,
the master starts at $t=0$ by waiting for the fastest single worker. Subsequently, let $t_k,~k=0,\ldots,n-1$, be the time when the master switches from waiting for the fastest $k$ workers to waiting for the fastest $k+1$ workers, with $t_0=0$.  For $t_{k-1}\leq t\leq t_k$, we have
\begin{equation}
\left \lvert \frac{d e(t)}{d t} \right \rvert = 
\frac{\alpha}{\mu_k} e^{ \frac{-\alpha(t-t_{k-1})}{\mu_k}} \left(e(t_{k-1})-\frac{\eta L \sigma^2}{2cks} \right). \label{eq7} 
\end{equation}
For $t_k\leq t \leq t_{k+1}$, we have
\begin{equation}
\left \lvert \frac{d e(t)}{d t} \right \rvert  = 
\frac{\alpha}{\mu_{k+1}} e^{ \frac{-\alpha(t-t_k)}{\mu_{k+1}}} \left(e(t_{k})-\frac{\eta L \sigma^2}{2c(k+1)s} \right). \label{eq8}
\end{equation}
The bound-optimal switching time $t_k$ is given by the smallest $t$ for which the rate in~\eqref{eq8} becomes larger than the rate in~\eqref{eq7}. Hence $t_k$ can be determined by evaluating the following,
\begin{equation}
\label{neq1}
\frac{\alpha}{\mu_{k+1}} \left(e(t_{k})-\frac{\eta L \sigma^2}{2c(k+1)s} \right) \geq \frac{\alpha}{\mu_k} e^{ \frac{-\alpha(t_k-t_{k-1})}{\mu_k}} \left(e(t_{k-1})-\frac{\eta L \sigma^2}{2cks} \right),
\end{equation}
where the LHS in~\eqref{neq1} is obtained by evaluating~\eqref{eq8} at $t=t_k$. The expression of $e(t_k)$ is given by
\begin{equation}
\label{neq2}
e(t_k) = \frac{\eta L \sigma^2}{2cks}+e^{ \frac{-\alpha (t_k-t_{k-1})}{\mu_k} } \left(e(t_{k-1})-\frac{\eta L \sigma^2}{2cks} \right). 
\end{equation}
Substituting \eqref{neq2} in \eqref{neq1} and simplifying we get
\begin{equation}
\label{neq3}
e^{ \frac{\alpha (t_k-t_{k-1})}{\mu_k} } \leq \frac{(\mu_{k+1}-\mu_k)(2csk(k+1)e(t_{k-1})-\eta L (k+1) \sigma^2)}{\eta L \sigma^2 \mu_k}.
\end{equation}
Taking logarithms on both sides of~\eqref{neq3} we obtain
\begin{equation}
t_k  = t_{k-1}+\frac{\mu_k}{\alpha} \bigg[ \ln \left(\mu_{k+1}-\mu_k \right) +\ln \left(2csk(k+1)e(t_{k-1})-\eta L (k+1) \sigma^2) \right) - \ln \left(\eta L \sigma^2\mu_k \right) \bigg] \label{eqq11}.
\end{equation} 
where $t_0=0$ and $e(t_0)=F(\mathbf{w}_0)-F^{\star}$.

\section{Acknowledgment}
This work was supported in part by the Army Research Lab (ARL) under Grant W911NF-1820181.

\section{Conclusion}
We considered a distributed computation setting in which a master wants to run a stochastic gradient descent algorithm with the help of $n$ workers. The master partitions the data in his possession among the workers and asks each worker to compute a partial gradient. We proposed adaptive fastest-$k$ SGD to mitigate the effect of stragglers and speed up the computation process while minimizing the download communication cost from the workers to the master. We showed that by waiting for a small number of workers, and increasing this number as the algorithm evolves, the master can \begin{enumerate*}[label=(\roman*)] \item increase the convergence rate as a function of the wall-clock time; \item reach a higher model accuracy; and \item reduce the communication cost \end{enumerate*}. We validated our theoretical findings with simulation results using linear regression on synthetic data and logistic regression on the MNIST dataset.

As for future research directions, one can consider methods that target reducing the upload communication cost, e.g., by using fewer workers per iteration and waiting for some stragglers. In addition, one can study the effect of combining the adaptive techniques introduced in this work with coding theoretic techniques in the setting where the data is distributed redundantly to the workers. Another direction for future work is to consider generalizations such as non-convex functions, variable step sizes, and alternative methods for detecting phase transitions in SGD. Furthermore, applying such ideas to the framework of federated learning could be also investigated.

\bibliography{Refs}
\bibliographystyle{ieeetr}

\appendix
\label{app}
\label{s2p3}
\noindent The result in Proposition~\ref{prop1} holds under the following conditions stated in~\cite{Gauri,Bottou}:
\begin{enumerate}

\item The loss function $F(\mathbf{w})$ is continuously differentiable and the gradient function of $F$, given by $\nabla F$, is Lipschitz continuous with Lipschitz constant $L > 0$, i.e.,
\begin{equation*}
\norm{\nabla F(\mathbf{w}_1)-\nabla F(\mathbf{w}_2)}_2 \leq L \norm{\mathbf{w}_1-\mathbf{w}_2}_2.
\end{equation*}
\item The loss function $F(\mathbf{w})$ is strongly convex, i.e., there exists a constant $c > 0$ such that
\begin{equation*}
2c(F(\mathbf{w})-F^{\star})\leq \norm{\nabla F(\mathbf{w})}_2^2 ~~\forall \ \mathbf{w},
\end{equation*}
where $F^{\star}=F(\mathbf{w}^{\star})$ is the loss function evaluated at the optimal point $\mathbf{w}^{\star}$.
{\color{blue}
}
\item The stochastic gradient is an unbiased estimate of the actual gradient and has bounded variance, i.e., for all iterations indexed by $j\in \N$,
\begin{align*}
\E \left[ \ghat_j \right] = \nabla F(\mathbf{w}_j), \ 
\Var [\ghat_j] \leq \dfrac{ \sigma^2}{s} + \dfrac{M_G}{s} \norm{\nabla F(\mathbf{w}_j) }_2^2,
\end{align*}
where the expectation is taken over the observed rows $\mathbf{a}_{\ell}$ at the master, and $\sigma$ and $M_G$ are positive scalars.
\end{enumerate}
\noindent The formal statement of the proposition is the following.
\addtocounter{proposition}{-1}
\begin{proposition}[formal]
\label{prop2}
Suppose that the loss function $F(\mathbf{w})$ is $c$-strongly convex and $L$-lipchitz, and consider a fixed learning rate $\eta$ such that $\eta\leq 1/[2L(\frac{M_G}{ks}+1)]$. Then, the error of fastest-$k$ SGD after $j$ iterations satisfies
\begin{equation*}
\mathbb{E}\left[F(\mathbf{w}_j)-F^{\star} \right]\leq \frac{\eta L \sigma^2}{2cks} +(1-\eta c)^{j}  \left(F(\mathbf{w}_0)-F^{\star}-\frac{\eta L \sigma^2}{2cks} \right) \label{eeq10}.
\end{equation*}
\end{proposition}

\begin{remark}
One can easily verify that the gradient estimator given in~\eqref{eeq6} satisfies the unbiasedness condition. Moreover, the bounded variance condition can be also easily shown when the optimization is done over a convex compact set $\mathcal{W}\subset \R^d$. Furthermore, the least square errors (L2) is an example of a loss function that satisfies the first and second conditions.
\end{remark}

\end{document}